\documentclass[sigconf]{acmart}
\usepackage{balance}
\usepackage{booktabs,makecell, multirow, tabularx}
\usepackage{xcolor}
\usepackage{hyperref}

\AtBeginDocument{%
  }

\copyrightyear{2024}
\acmYear{2024}
\setcopyright{acmlicensed}
\acmConference[KDD '24] {Proceedings of the 30th ACM SIGKDD Conference on Knowledge Discovery and Data Mining }{August 25--29, 2024}{Barcelona, Spain.}
\acmBooktitle{Proceedings of the 30th ACM SIGKDD Conference on Knowledge Discovery and Data Mining (KDD '24), August 25--29, 2024, Barcelona, Spain}
\acmISBN{979-8-4007-0490-1/24/08}
\acmDOI{10.1145/3637528.3671869}

\begin{document}
\begin{sloppypar}

\title{Conditional Logical Message Passing Transformer for Complex Query Answering}

\author{Chongzhi Zhang}
\orcid{0009-0000-0515-4795}
\affiliation{%
  \institution{South China University of Technology}
  \city{Guangzhou}
  \country{China}
}
\email{cschongzhizhang@mail.scut.edu.cn}

\author{Zhiping Peng}
\orcid{0000-0002-4561-2778}
\affiliation{%
  \institution{Guangdong University of Petrochemical Technology}
  \city{Maoming}
  \country{China}}
\affiliation{%
  \institution{Jiangmen Polytechnic}
  \city{Jiangmen}
  \country{China}}
\email{pengzp@foxmail.com}

\author{Junhao Zheng}
\orcid{0000-0001-9124-2467}
\affiliation{%
  \institution{South China University of Technology}
  \city{Guangzhou}
  \country{China}
}
\email{junhaozheng47@outlook.com}

\author{Qianli Ma}
\orcid{0000-0002-9356-2883}
\authornote{Corresponding author}
\affiliation{%
  \institution{South China University of Technology}
  \city{Guangzhou}
  \country{China}
}
\email{qianlima@scut.edu.cn}


\begin{abstract}
  Complex Query Answering (CQA) over Knowledge Graphs (KGs) is a challenging task. Given that KGs are usually incomplete, neural models are proposed to solve CQA by performing multi-hop logical reasoning. However, most of them cannot perform well on both one-hop and multi-hop queries simultaneously. Recent work proposes a logical message passing mechanism based on the pre-trained neural link predictors. While effective on both one-hop and multi-hop queries, it ignores the difference between the constant and variable nodes in a query graph. In addition, during the node embedding update stage, this mechanism cannot dynamically measure the importance of different messages, and whether it can capture the implicit logical dependencies related to a node and received messages remains unclear. In this paper, we propose Conditional Logical Message Passing Transformer (CLMPT), which considers the difference between constants and variables in the case of using pre-trained neural link predictors and performs message passing conditionally on the node type. We empirically verified that this approach can reduce computational costs without affecting performance. Furthermore, CLMPT uses the transformer to aggregate received messages and update the corresponding node embedding. Through the self-attention mechanism, CLMPT can assign adaptive weights to elements in an input set consisting of received messages and the corresponding node and explicitly model logical dependencies between various elements. Experimental results show that CLMPT is a new state-of-the-art neural CQA model. \href{https://github.com/qianlima-lab/CLMPT}{https://github.com/qianlima-lab/CLMPT}. 
\end{abstract}





\maketitle

\section{Introduction}
\label{Introduction}

Knowledge graphs (KGs) store factual knowledge in the form of triples that can be utilized to support a variety of downstream tasks \cite{saxena2020improving, wang2021learning, xiong2017explicit}. 
However, given that modern KGs are usually auto-generated \cite{toutanova2015observed} or built through crowd-sourcing \cite{vrandevcic2014wikidata}, 
so real-world KGs \cite{bollacker2008freebase, suchanek2007yago, carlson2010toward} are often considered noisy and incomplete, which is also known as the Open World Assumption \cite{libkin2009open,ji2021survey}. 
Answering queries on such incomplete KGs is a fundamental yet challenging task. 
To alleviate the incompleteness of KGs, knowledge graph representation methods \cite{bordes2013translating, trouillon2016complex, sun2018rotate, dettmers2018convolutional}, which can be viewed as neural link predictors \cite{arakelyan2020complex}, have been developed. 
They learn representations based on the available triples and generalize them to unseen triples. 
Such neural link predictors can answer one-hop atomic queries on incomplete KGs. 
But for multi-hop complex queries, the query-answering models need to perform multi-hop logical reasoning over incomplete KGs. That is, not only to utilize available knowledge to predict the unseen one but also execute logical operators, such as conjunction ($\wedge$), disjunction ($\vee$), and negation ($\neg$) \cite{ren2020beta, wang2022benchmarking}.

\begin{figure}
    \centering
    \includegraphics[width=1\linewidth]{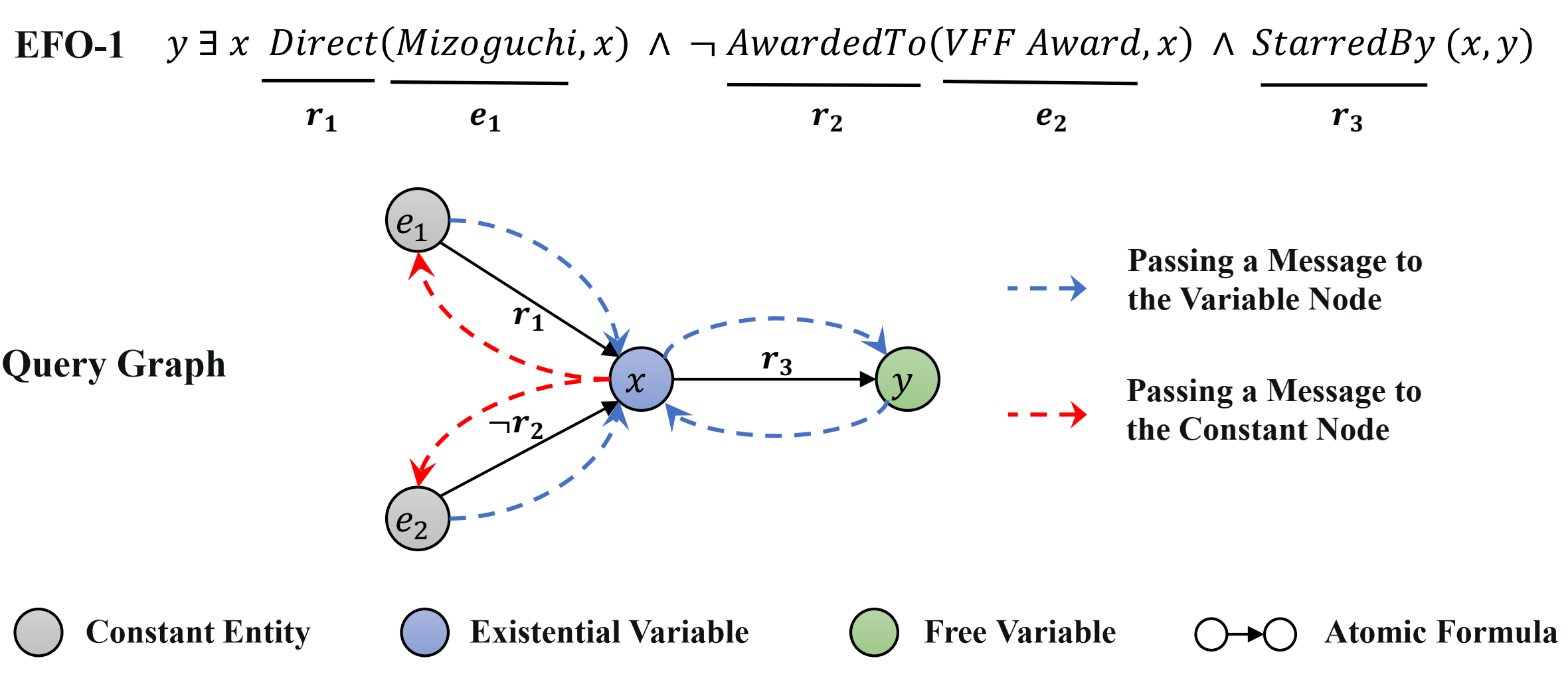}
    \caption{EFO-1 query and its query graph for the question “Who starred the films that were directed by Mizoguchi but never won a Venice Film Festival Award?”. }
    \label{figure1}
\end{figure}

Recently, neural models \cite{ren2020beta, amayuelas2022neural, zhang2021cone,bai2023knowledge} have been proposed to solve Complex Query Answering (CQA) over incomplete KGs. 
The complex queries that these models aim to address belong to an important subset of the first-order queries, specifically Existentially quantified First Order queries with a single free variable (EFO-1) \cite{wang2022benchmarking, wang2023logical}. 
Such EFO-1 queries contain logical operators, including conjunction, disjunction, and negation, as well as the existential quantifier ($\exists$) \cite{ren2020beta}. 
As shown in Figure \ref{figure1}, one can get the corresponding EFO-1 formula and query graph given a question. 
The query graph is a graphical representation of the EFO-1 query, where each edge is an atomic formula that contains a predicate with a (possible) negation operator, 
and each node represents an input constant entity or a variable. 
Most neural CQA models convert the query graph into the computation graph, where logical operators are replaced with corresponding set operators. 
These models embed the entity set into specific vector spaces \cite{ren2020beta, zhang2021cone, chen2022fuzzy, yang2022gammae} and execute set operations parameterized by the neural networks according to the computational graph. 
However, these models cannot perform well on both one-hop and multi-hop queries simultaneously. 
They tend to be less effective than classical neural link predictors \cite{trouillon2016complex} on one-hop atomic queries \cite{arakelyan2020complex, wang2023logical}. 

To this end, Logical Message Passing Neural Network (LMPNN) \cite{wang2023logical} proposes a message passing framework based on pre-trained neural link predictors. Specifically, for each edge in the query graph, LMPNN uses the pre-trained neural link predictor to infer an intermediate embedding for a node given neighborhood information. The intermediate embedding can be interpreted as a logical message passed by the neighbor on the edge. Figure \ref{figure1} demonstrates the logical message passing with red/blue arrows. 
Then, LMPNN aggregates messages and updates node embeddings in a manner similar to Graph Isomorphism Network (GIN) \cite{xu2018how}. 
While effective on both one-hop and multi-hop queries, LMPNN ignores the difference between constant and variable nodes.  
Logical message passing can be viewed as learning an embedding for the variable  by utilizing information of constant entities. 
It makes this embedding similar or close to the embedding of the solution entity in the embedding space of the pre-trained neural link predictor. 
However, as shown by the red arrows in Figure \ref{figure1}, LMPNN also passes messages to constant entitiy nodes with pre-trained embeddings and updates their embeddings, which we believe is unnecessary and experimentally demonstrated. 
In addition, for node embedding updating, despite the expressive power of GIN in graph structure learning \cite{xu2018how}, it cannot dynamically measure the importance of different logical messages. 
Furthermore, it is unclear whether the GIN-like approach can capture the implicit complex logical dependencies between messages received by a node and between messages and that node. 
Explicitly modeling the above implicit logical dependencies while being able to measure the importance of different messages dynamically is an open challenge we take on in this paper. 

We propose Conditional Logical Message Passing Transformer (CLMPT), a special Graph Neural Network (GNN) for CQA. 
Each CLMPT layer has two stages: 
(1) passing the logical message of each node to all of its neighbors that are not constant entity nodes; 
(2) updating the embedding of each variable node based on the logical messages received by the variable node. 
This means that during forward message passing, CLMPT does not use the pre-trained neural link predictor to infer the intermediate embeddings (i.e. logical messages) for constant entity nodes with pre-trained embeddings and does not update constant node embeddings. 
We call such a mechanism Conditional Logical Message Passing, 
where the condition depends on whether the node type is constant or variable. 
We empirically verified that it can reduce computational costs without affecting performance. 
Furthermore, 
CLMPT uses the transformer \cite{vaswani2017attention} to aggregate messages and update the corresponding node embedding. 
The self-attention mechanism in the transformer can dynamically measure the importance of different elements in the input and capture the interactions between any two elements. It allows us to explicitly model the logical dependencies between messages received by a node and between messages and that node. 
We conducted experiments on three popular KGs: FB15k \cite{bordes2013translating}, FB15k-237 \cite{toutanova2015observed}, and NELL995 \cite{xiong2017deeppath}. 
The experimental results show that CLMPT can achieve a strong performance. 

The main contributions of this paper are summarized as follows: 
\begin{itemize}
    \item We propose a conditional logical message passing mechanism, which can effectively reduce the computational costs and even improve the model performance to some extent. 
    \item 
    We propose a transformer-based node embedding update scheme. It can dynamically measure the importance of different intermediate embeddings of a variable node and capture possible implicit dependencies between these embeddings. 
    \item Extensive experiments results on several benchmark datasets show that CLMPT is a new state-of-the-art neural CQA model. 
\end{itemize}

\section{Related Work}
\label{Related Work}

\paragraph{Neural Link Predictors}

Reasoning over KGs with missing knowledge is one of the fundamental problems in Artificial Intelligence and has been widely studied. 
Traditional KG reasoning tasks such as link prediction \cite{getoor2007introduction, trouillon2016complex} are essentially one-hop atomic query problems. 
Representative methods for link prediction are knowledge graph representation methods \cite{bordes2013translating, nickel2011three, yang2015embedding, sun2018rotate, trouillon2016complex, dettmers2018convolutional}, which can be regarded as neural link predictors \cite{arakelyan2020complex}. 
Specifically, they embed entities and relations into continuous vector spaces and predict unseen triples by scoring triples with a well-defined scoring function. 
Such latent feature models can effectively answer one-hop atomic queries over incomplete KGs. 
Other methods for link prediction include rule learning \cite{sadeghian2019drum, zhang2019iteratively}, text representation learning \cite{wang2021kepler, saxena2022sequence, wang2021structure}, and GNNs \cite{zhang2022rethinking, zhang2022knowledge,Vashishth2020Composition-based}.

\paragraph{Neural Complex Query Answering}

Complex queries over KGs can be regarded as one-hop atomic queries combined with existential first-order logic operators. 
The scope of complex queries that existing works can answer is expanded from conjunctive queries \cite{hamilton2018embedding, kotnis2021answering}, to Existential Positive First-Order (EPFO) queries \cite{Ren*2020Query2box:,arakelyan2020complex}, and more recently to the existential first-order \cite{ren2020beta}, or more specifically to EFO-1 \cite{wang2022benchmarking, wang2023logical} and
$\text{EFO}_{k} $ queries \cite{yin2023rm}. 
Recently, neural query embedding (QE) models have been proposed to answer complex queries.  
Most QE models embed entity sets into specific continuous vector spaces with various forms, including geometric shapes \cite{choudhary2021self,amayuelas2022neural,hamilton2018embedding,Ren*2020Query2box:,zhang2021cone,liu2021neural,nguyen2023cyle,bai2022query2particles}, probabilistic distributions \cite{ren2020beta, yang2022gammae,choudhary2021probabilistic}, fuzzy logic \cite{chen2022fuzzy}, and bounded histograms on uniform grids \cite{wang2023wasserstein}. 
They represent the complex query in the form of the computation graph and perform neural set operations in the corresponding embedding spaces according to the computation graph. Such neural set operations are usually modeled by complex neural networks \cite{tolstikhin2021mlp, vaswani2017attention}. 
In spite of getting query embedding with neural set operations, other QE models represent the queries as graphs to embed queries with GNNs \cite{daza2020message, alivanistos2022query} or Transformers \cite{kotnis2021answering, liu2022mask, wang2023query,bai2023sequential}. 
However, most of these graph representation learning models only focus on EPFO queries, and how they handle the logical negation operator is unclear. 
Moreover, the aforementioned neural models are still less effective than a simple neural link predictor \cite{trouillon2016complex} on one-hop atomic queries.

There is also a class of neural CQA models based on pre-trained neural link predictors. 
CQD-CO \cite{arakelyan2020complex} proposes a continuous optimization strategy that uses pre-trained neural link predictors \cite{trouillon2016complex} and logical t-norms \cite{klement2013triangular} to solve complex queries. 
However, according to \cite{wang2023logical}, CQD-CO performs badly on negative queries. 
Q2T \cite{xu2023query2triple} is based on graph transformers \cite{ying2021transformers} and pre-trained neural link predictors to encode complex queries. 
LMPNN \cite{wang2023logical}, which is most related to our work, proposes a logical message passing framework based on pre-trained neural link predictors. 
Through one-hop inference on atomic formulas, 
LMPNN performs bidirectional logical message passing on each edge in the query graph. 
However, such a mechanism ignores the difference between the constant and variable nodes. 
The essence of logical message passing lies in leveraging the pre-trained entity embeddings corresponding to constant nodes to infer variable node embeddings. It aligns the variable node embeddings more closely to the embeddings of solution entities in the embedding space of the pre-trained neural link predictor. It is important to note that using the intermediate states of the variable nodes to update the embeddings of constant nodes can introduce noise. 
This occurs because constant nodes possess stable, pre-trained embeddings corresponding to specific entities, which do not require updates via message passing. 
Therefore, updating constant node embeddings with information from variable nodes disrupts the integrity of these established embeddings. 
In addition, at the node level, LMPNN aggregates the received logical messages and updates the node embeddings in a manner similar to GIN \cite{xu2018how}. 
Despite the expressive power of GIN in graph structure learning, it cannot dynamically measure the importance of different messages. 
Whether such a GIN-like approach can capture the implicit logical dependencies between messages received by a node and between messages and that node remains unclear. 
By contrast, our work only performs conditional message passing on the query graph, 
which avoids unnecessary computation costs. 
Furthermore, 
by using the transformer \cite{vaswani2017attention} to aggregate the message and update the corresponding node embedding, 
our work can assign adaptive weights to elements in an input set consisting of received messages and the corresponding node and explicitly model logical dependencies between various elements.

\paragraph{Symbolic Integration Models}
\label{relate work sym}

In addition to neural CQA models, some recent studies integrate symbolic information into neural models, which can be called symbolic integration models, also known as neural-symbolic models \cite{arakelyan2020complex,xu2022neural,zhu2022neural,arakelyan2023adapting,bai2023answering}. 
These models estimate the probability of whether each entity is the answer at each intermediate step in the process of multi-hop logical reasoning. 
This makes the size of the intermediate states for symbolic reasoning in symbolic integration models scale linearly with the size of the KG. 
Therefore, compared with neural CQA models where the intermediate embeddings are of fixed dimension, symbolic integration models require more computing resources and are more likely to suffer from scalability problems on large-scale KGs \cite{ren2023neural, wang2023logical}.

\section{Background}
\label{Background}

\paragraph{Model-theoretic Concepts for Knowledge Graphs}

A Knowledge Graph $\mathcal{KG}$ consists of a set of entities $\mathcal{V}$ and a set of relations $\mathcal{R}$. It can be defined as a set of triples $\mathcal{E} ={(e_{h_{i}},r_{i},e_{t_{i}})} \subseteq \mathcal{V} \times \mathcal{R} \times \mathcal{V}$, 
namely $\mathcal{KG}=(\mathcal{V}, \mathcal{E}, \mathcal{R})$. 
A first-order language $\mathcal{L}$ can be defined as $(\mathcal{F},\mathcal{R},\mathcal{C})$, where $\mathcal{F}$, $\mathcal{R}$, and $\mathcal{C}$ are sets of symbols for functions, relations, and constants, respectively \cite{wang2023logical}. 
Under language $\mathcal{L_{KG}}$, the $\mathcal{KG}$ can be represented as a first-order logic knowledge base \cite{arakelyan2020complex}. 
In this case, the relation symbols in $\mathcal{R}$ denote binary relations, the constant symbols in $\mathcal{C}$ represent the entities, and the function symbol set satisfies $\mathcal{F}=\emptyset$. 
In other words, $\mathcal{KG}$ is an $\mathcal{L_{KG}}$-structure, 
where each entity $e\in \mathcal{V}$ is also a constant $c\in \mathcal{C}=\mathcal{V}$ and each relation $r \in \mathcal{R}$ is a set $r \subseteq \mathcal{V} \times \mathcal{V}$. We have $r(t_{1}, t_{2}) = True$ when $(t_{1}, t_{2}) \in r$. 
An atomic formula is either $r(t_{1}, t_{2})$ or $\neg r(t_{1}, t_{2})$, where $t_{i}$ is a term that can be a constant or a variable, and $r$ is a relation that can be viewed as a binary predicate \cite{arakelyan2020complex}. 
A variable is a bound variable when associated with a quantifier. Otherwise, it is a free variable. 
By adding connectives (conjunction $\wedge$, disjunction $\vee$, and negation $\neg$) to such atomic formulas and adding quantifiers (existential $\exists$ and universal $\forall$) to variables, we can inductively define the first order formula \cite{wang2023logical,marker2006model}. 

\paragraph{EFO-1 Queries}

In this paper, we consider Existential First Order queries with a single free variable (EFO-1) \cite{wang2022benchmarking, wang2023logical}. 
Such EFO-1 queries are an important subset of first-order queries, using existential quantifier, conjunction, disjunction, and atomic negation. 
Following the previous studies \cite{ren2020beta, Ren*2020Query2box:, wang2023logical}, 
we define the EFO-1 query as the first order formula in the following Disjunctive Normal Form (DNF) \cite{davey2002introduction}: 
\begin{equation}
\begin{split}
    q[y,x_{1},...,x_{k}] = \exists x_{1},...,\exists x_{k} (a_{1}^{1}\wedge ...\wedge a_{n_{1}}^{1}) \\
    \vee... \vee (a_{1}^{d}\wedge ...\wedge a_{n_{d}}^{d}), \label{first-order formula}
\end{split}
\end{equation}
where $y$ is the only free variable, $ x_{1},...,x_{k}$ are existential variables. 
$a_{j}^{i}$ are atomic formulas that can be either negated or not: 
\begin{equation}
\begin{split}
     a_{j}^{i}=\begin{array}{l} 
  \left\{\begin{matrix} 
  r(e_{c},v)\ or\ r(v,v') \\ 
  \neg r(e_{c},v)\ or\ \neg r(v,v') 
\end{matrix}\right.    
\end{array}
, \label{atomic formula}
\end{split}
\end{equation}
where $e_{c} \in \mathcal{V}$ is an input constant entity, $v,v'\in \{y,x_{1},...,x_{k}\}$ are variables, $v\ne v'$. 
To address the EFO-1 query $q$, the corresponding answer entity set $\llbracket q \rrbracket \subseteq \mathcal{V}$ needs to be determined, 
where $\llbracket q \rrbracket$ is a set of entities such that $e \in \llbracket q \rrbracket$ iff $q[y=e,x_{1},...,x_{k}]=True$. 
In particular, since we use DNF to represent the first order formula, the Equation \ref{first-order formula} can also be expressed as the disjunction of conjunctive queries: 
\begin{equation}
\begin{split}
    q[y,x_{1},...,x_{k}] = CQ_{1}(y,x_{1},...,x_{k})    \vee... \vee CQ_{d}(y,x_{1},...,x_{k}), \label{DNF fol formula}
\end{split}
\end{equation}
where $CQ_{i} = \exists x_{1},...,\exists x_{k}\ a_{1}^{i}\wedge ...\wedge a_{n_{i}}^{i}$ is a conjunctive query. 
According to \cite{Ren*2020Query2box:, wang2023logical}, we can get the answer set $\llbracket q \rrbracket$ by taking the union of the answer sets of each conjunctive query, namely $\llbracket q \rrbracket =  {\textstyle \bigcup_{i=1}^{d}} \llbracket CQ_{i} \rrbracket$. 
This means that solving all conjunctive queries $CQ_{i},1 \le i\le d$ yields the answer set $\llbracket q \rrbracket$ for the EFO-1 query $q$. 
Such DNF-based processing can solve complex queries involving disjunction operators in a scalable way \cite{Ren*2020Query2box:, ren2023neural,ren2022smore}.

\paragraph{Query Graph}

Since disjunctive queries can be solved in a scalable manner by transforming queries into the disjunctive normal form, it is only necessary to define query graphs for conjunctive queries. 
We follow previous works \cite{wang2023logical,Ren*2020Query2box:} and represent the conjunctive query as the query graph where the terms are represented as nodes connected by the atomic formulas. 
That is, each edge in the query graph is an atomic formula. 
According to the definition in Equation \ref{atomic formula}, an atomic formula contains relations, negation information, and terms. 
Therefore, each node in the query graph is either a constant entity or a free or existential variable, as illustrated in Figure \ref{figure1}. 
In this paper, the term "constants" has specific meanings depending on the context. When discussing an EFO-1 query, "constants" refers to the input constant entities within that query. In the context of the query graph, "constants" denotes the constant nodes that correspond to these input constant entities.

\paragraph{Neural Link Predictors}

A neural link predictor has a corresponding scoring function that is used to learn the embeddings of entities and relations in KG. 
For a triple, the embeddings of the head entity, relation, and tail entity are $h$, $r$, and $t$, respectively. 
The scoring function $\phi (h,r,t)$ can calculate the likelihood score of whether the triple exists. 
By using the scoring function $\phi (h,r,t)$ and the sigmoid function $\sigma$, neural link predictors can produce a continuous truth value $\psi (h,r,t)\in [0,1]$ for a triple. 
For example, the scoring function of ComplEx \cite{trouillon2016complex} is as follows: 
\begin{equation}
    \phi (h,r,t)=Re \left (  \left \langle h\otimes r,\overline{t}  \right \rangle\right ), \\\label{scoringfunction}
\end{equation}
where $\otimes$ represents the element-wise complex number multiplication, $\left \langle \cdot , \cdot  \right \rangle$ is the complex inner product, and $Re$ means taking the real part of a complex number. Based on $\phi (h,r,t)$, we can obtain the corresponding truth value: 
\begin{equation}
    \psi(h,r,t)=\sigma (\phi (h,r,t)) \\\label{truthvalue}
    .
\end{equation}
In this work, the neural link predictor handles not only specific entity embeddings but also variable embeddings. 
That is, the embeddings $h,t$ correspond to embedding of terms, which can be the embeddings of constant entities or the embeddings of variables. 
Following previous works \cite{arakelyan2020complex,wang2023logical}, in our work, we use ComplEx-N3 \cite{trouillon2016complex,lacroix2018canonical} as the neural link predictor of CLMPT.

\paragraph{One-hop Inference on Atomic Formulas}

As shown in Figure \ref{figure1}, each edge in a query graph is an atomic formula containing the information of relation, logical negation, and terms. 
Passing a logical message on an edge is essentially an operation of utilizing this information to perform one-hop inference on an atomic formula. 
On each edge, when a node is at the head position, its neighbor is at the tail position, and vice versa. 
A logical message passed from a neighbor to a node is essentially an intermediate embedding of that node. 
Prior work \cite{wang2023logical} proposes to obtain the intermediate embeddings of nodes by one-hop inference that maximizes the continuous truth value of the (negated) atomic formulas. 
Specifically, a logical message encoding function $\rho$ is proposed to perform one-hop inference. 
Such a function has four input parameters, including neighbor embedding, relation embedding, direction information ($h\to t$ or $t\to h$), and logical negation information ($0$ for no negation and $1$ for with negation), and it has four cases depending on the input parameters. 
Given the tail embedding $t$ and relation embedding $r$ on a non-negated atomic formula, 
$\rho$ is formulated in the form of continuous truth value maximization to infer the head embedding $\hat{h}$: 
\begin{equation}
\begin{split}
    \hat{h} =\rho (t,r,t\to h,0):=\underset{x\in \mathcal{D} }{arg\ max}\ \psi(x,r,t), \label{rho1}
\end{split}
\end{equation}
where $\mathcal{D}$ is the search domain for the embedding $x$. 
Similarly, the tail embedding $\hat{t}$ on a non-negated atomic formula can be inferred given $h$ and $r$: 
\begin{equation}
\begin{split}
    \hat{t} =\rho (h,r,h\to t,0):=\underset{x\in \mathcal{D} }{arg\ max}\ \psi(h,r,x). \label{rho2}
\end{split}
\end{equation}
According to the fuzzy logic negator \cite{hajek2013metamathematics}, one has $\psi(h,\neg r,t)=1-\psi(h,r,t)$. Therefore, 
the estimation of head and tail embeddings on negated atomic formulas can be defined as follows: 
\begin{equation}
\begin{split}
    \hat{h} =\rho (t,r,t\to h,1) := \underset{x\in \mathcal{D} }{arg\ max}\ [1 - \psi(x,r,t)]
    , \label{rho3}
\end{split}
\end{equation}
\begin{equation}
\begin{split}
    \hat{t} =\rho (h,r,h\to t,1) := \underset{x\in \mathcal{D} }{arg\ max}\ [1 - \psi(h,r,x)]. \label{rho4}
\end{split}
\end{equation}
For the situation where ComplEx-N3 \cite{lacroix2018canonical,trouillon2016complex} is selected as the neural link predictor, according to the propositions and proofs in \cite{wang2023logical}, the logical message encoding function $\rho$ has the following closed-form with respect to the complex embedding $r$ and $t$ when inferring the head embedding $\hat{h}$ on a non-negated atomic formula: 
\begin{equation}
\begin{split}
    \rho (t,r,t\to h,0) &=\underset{x\in \mathbb{C}^{d} } {arg\ max}\ \left \{ Re \left (  \left \langle \overline{r}\otimes t,\overline{x}  \right \rangle\right )-\lambda {\left \| x \right \| }^{3} \right \} \\
    &= \frac{\overline{r}\otimes t}{\sqrt{3\lambda \left \| \overline{r}\otimes t \right \|  } },  \label{rho11}
\end{split}
\end{equation}
Similarly, for the other three cases, the encoding functions $\rho$ are as follows: 
\begin{equation}
\begin{split}
    \rho (h,r,h\to t,0) = \frac{r\otimes h}{\sqrt{3\lambda \left \| r\otimes h \right \|  } },  \label{rho21}
\end{split}
\end{equation}
\begin{equation}
\begin{split}
    \rho (t,r,t\to h,1) = \frac{-\overline{r}\otimes t}{\sqrt{3\lambda \left \| \overline{r}\otimes t \right \|  } },  \label{rho31}
\end{split}
\end{equation}
\begin{equation}
\begin{split}
    \rho (h,r,h\to t,1) = \frac{-r \otimes h}{\sqrt{3\lambda \left \| r\otimes h \right \|  } }.  \label{rho41}
\end{split}
\end{equation}
For Equations \ref{rho11}--\ref{rho41}, $\lambda$ is a hyperparameter. 
In our work, we follow prior work \cite{wang2023logical} and use the above logical message encoding function $\rho$ to compute messages passed to the variable nodes.

\begin{figure*}
    \centering
    \includegraphics[width=1\textwidth]{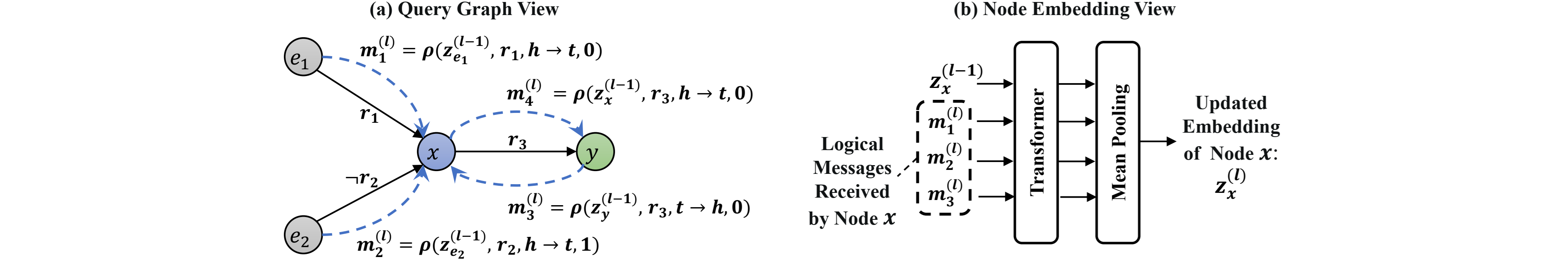}
    \caption{An illustration of the two-stage procedures of CLMPT at the $l$-th layer. 
    (a) Passing the logical message to the variable node in the neighborhood. (b) Updating the existential variable node embedding with the received messages and a transformer encoder. Similarly, such a node embedding update scheme is applied to other variable nodes, such as the free variable node $y$. }
    \label{figure2}
\end{figure*}

\section{Proposed Method}

In this section, we propose Conditional Logical Message Passing Transformer (CLMPT) to answer complex queries. 
As a special message passing neural network \cite{gilmer2017neural}, each CLMPT layer has two stages: 
(1) passing the logical message of each node to all of its neighbors that are not constant entity nodes; 
(2) updating the embedding of each variable node based on the messages received by the variable node. 
Figure \ref{figure2} illustrates how these two stages work using the query graph in Figure \ref{figure1}. 
Then, we can use the final layer embedding of the free variable node to predict the answer entities to the corresponding query. 
In the following subsections, we first describe how each layer of CLMPT performs conditional logical message passing and updates the variable node embeddings. 
After that, we represent the training objective of our method. 
Finally, we introduce how to answer complex queries with CLMPT.

\subsection{Conditional Logical Message Passing}

A query graph contains two types of nodes: constant entity nodes and variable nodes. 
We decide whether to pass logical messages to a node based on its type. 
Specifically, for any neighbor of a node, when the neighbor is a variable node, we use the logical message encoding function $\rho$ to calculate the corresponding message and pass it to the neighbor; when the neighbor is a constant entity node, we do not pass the logical message to it. Figure \ref{figure2} (a) demonstrates the conditional logical message passing with blue arrows. Every node in the query graph except the constant entity nodes receives the logical message from all of its neighbors.

\subsection{Node Embedding Conditional Update Scheme}

For a constant entity node, we denote its embedding at the $l$-th layer as $z_{e}^{(l)}$ and let $z_{v}^{(l)}$ be the embedding of a variable node at the $l$-th layer. Next, we discuss how to calculate the $z_{e}^{(l)}$ and $z_{v}^{(l)}$ from the input layer $l = 0$ to latent layers $l > 0$. 
For a constant entity node, $z_{e}^{(0)}$ is the corresponding entity embedding in the pre-trained neural link predictor, which can be frozen or optimized continuously. 
For $z_{v}^{(0)}$, we follow previous works \cite{wang2023logical} and assign two learnable embeddings $v_{x}, v_{y}$ for the existential variable node $x$ and the free variable node $y$ respectively, 
and set that all existential variable nodes $x_{i}$ share one $v_{x}$, 
namely $z_{x_{i}}^{(0)}=v_{x}$ and $z_{y}^{(0)}=v_{y}$. 
Similar to conditional logical message passing, for updating the node embeddings, we also do not consider constant entity nodes. 
That is, we only update node embeddings for variable nodes in the query graph. 
Thus, the embeddings of the constant entity nodes are the same at each layer, 
namely $z_{e}^{(l)}=z_{e}^{(0)}$. 
While for $z_{v}^{(l)}$, we use the corresponding information from the $(l-1)$-th layer to update it. 
Specifically, for a variable node $v$, we represent its neighbor set in the query graph as $\mathcal{N} (v)$. 
For each neighbor node $n \in \mathcal{N} (v)$, one can obtain information about the edge (i.e., the atomic formula) between $n$ and $v$, which contains the neighbor embedding $z_{n}^{(l-1)} \in \mathcal{D}$, the relation $r_{nv} \in \mathcal{R}$, the direction $D_{nv} \in \{ h\to t,t\to h\}$, and the negation indicator $Neg_{nv} \in \{0,1\}$. 
Based on the edge information, we can use the logical message encoding function $\rho$ to compute the logical message $m^{(l)}$ that $n$ passes to $v$: 
\begin{equation}
\begin{split}
    m^{(l)} =\rho (z_{n}^{(l-1)},r_{nv},D_{nv},Neg_{nv}). \label{rho-nv}
\end{split}
\end{equation}
Let $k_{v}$ be the number of neighbor nodes in $\mathcal{N} (v)$, where $k_{v} \ge 1$. 
For the variable node $v$, it receives $k_{v}$ logical messages from $k_{v}$ neighbors: $ m^{(l)}_{1},...,m^{(l)}_{k_{v}}$. 
We use the transformer \cite{vaswani2017attention} to encode the input set consisting of these logical messages and the variable node embedding $z_{v}^{(l-1)}$ to obtain the updated embedding of node $v$: 
\begin{equation}
\begin{split}
    z_{v}^{(l)}=Mean(TE^{(l)}(m^{(l)}_{1},...,m^{(l)}_{k_{v}},z_{v}^{(l-1)})), \label{trme}
\end{split}
\end{equation}
where $Mean$ is a mean pooling layer. 
$TE$ is a standard transformer encoder stacked with multiple layers of transformer encoder blocks, where the attention computation approach is consistent with the bidirectional multi-head self-attention of the vanilla transformer \cite{vaswani2017attention}. 
Specifically, given an input embedding set $X$, it is first projected to query ($Q$), key ($K$), and value ($V$) matrices through a linear projection such that $Q=XW^Q$, $K=XW^K$ and $V=XW^V$ respectively. Then, the self-attention can be compute via 
\begin{equation}
\begin{split}
    Attention(Q,K,V)=softmax(\frac{QK^T}{\sqrt{d_K} } )V , \label{attention}
\end{split}
\end{equation}
where $d_K$ represents the dimension of $K$, and $W^Q$, $W^K$, $W^V$ are parameter matrices. 
We use multi-head attention, which concatenates multiple instances of Equation \ref{attention} and then obtains the output through a linear projection. 
It is worth noting that since the input to Equation \ref{trme} is a set consisting of the node embedding and the messages rather than a sequence, we do not use any positional encoding. This means that our architecture is permutation invariant. 
Through the self-attention mechanism, 
we can explicitly model the complex logical dependencies between various elements in the input set and dynamically measure the importance of different elements. 
Figure \ref{figure2} (b) shows how to obtain the updated embedding of the existential variable node $x$ through the encoding process defined by Equation \ref{trme}.

\subsection{Training Objective}

We use the Noisy Contrastive Estimation (NCE) loss proposed in \cite{ma2018noise} to train CLMPT. 
Specifically, we denote the positive data as $\{(a_{i},q_{i})\}_{i=1}^{N}$, where $a_{i} \in \llbracket q_{i} \rrbracket$. 
Our optimization involves $K$ uniformly sampled noisy answers from the entity set. 
The training objective is as follows: 
\begin{equation}
\begin{split}
    L_{NCE}=-\frac{1}{N} \sum_{i=1}^{N} log\left [\frac{F(z_{a_{i}},z(q_{i}),T)}{F(z_{a_{i}},z(q_{i}),T)+\sum_{k=1}^{K}F(z_{n_{ik}},z(q_{i}),T) } \right ], \label{nce1}
\end{split}
\end{equation}
\begin{equation}
\begin{split}
    F(x,y,z)=exp[cos(x,y)/z], \label{nce2}
\end{split}
\end{equation}
where $cos(\cdot,\cdot)$ is the cosine similarity, $T$ is a hyperparameter, $z_{a_{i}}$ is the embedding of positive answer $a_{i}$, $z_{n_{ik}}$ is the embedding of the $k$-th noisy samples, 
and $z(q_{i})$ is the embedding of the predicted answer to the query $q_{i}$, corresponding to the embedding of the free variable of the query $q_{i}$ at the final layer of CLMPT.

\subsection{Answering Complex Queries with CLMPT}
\label{4.4}

For a complex query defined in Equation \ref{DNF fol formula}, i.e., a DNF query, we follow the previous works \cite{ren2020beta, Ren*2020Query2box:} and use DNF-based processing to get answer entities. 
Specifically, we estimate the predicted answer embeddings for each sub-conjunctive query of the complex query. Then, the answer entities are ranked by the maximal cosine similarity against these predicted answer embeddings. Therefore, CLMPT only needs to consider the query graph of the conjunctive query.

For a given conjunctive query $q$, let the depth of CLMPT be $L$, and we apply the CLMPT layers $L$ times to the query graph of $q$. Then, we can obtain the free variable embedding $z_{y}^{(L)}$ at the final layer. We use $z_{y}^{(L)}$ as the embedding of the predicted answer entity, namely $z(q)=z_{y}^{(L)}$. Based on the cosine similarity between $z(q)$ and the entity embeddings in the pre-trained neural link predictor, we can rank the entities to retrieve answers. 
For the depth $L$, 
according to the analyses in \cite{wang2023logical}, $L$ should be the largest distance between the constant entity nodes and the free variable node to ensure the free variable node successfully receives all logical messages from the constant entity nodes.  
This means $L$ is not determined, and $L$ should change dynamically with different conjunctive query types. 
Thus, we follow prior work \cite{wang2023logical} and assume all $L$ CLMPT layers share the same transformer encoder.

In this case, the parameters in CLMPT include two embeddings for existential and free variables, a transformer encoder, and the parameters in the pre-trained neural link predictor. 
For the pre-trained neural link predictor, it can be frozen or not. 
In our work, we continue to optimize pre-trained neural link predictors by default, that is, we do not freeze pre-trained embeddings of entities and relations. 

\begin{figure}
    \centering
    \includegraphics[width=0.95\linewidth]{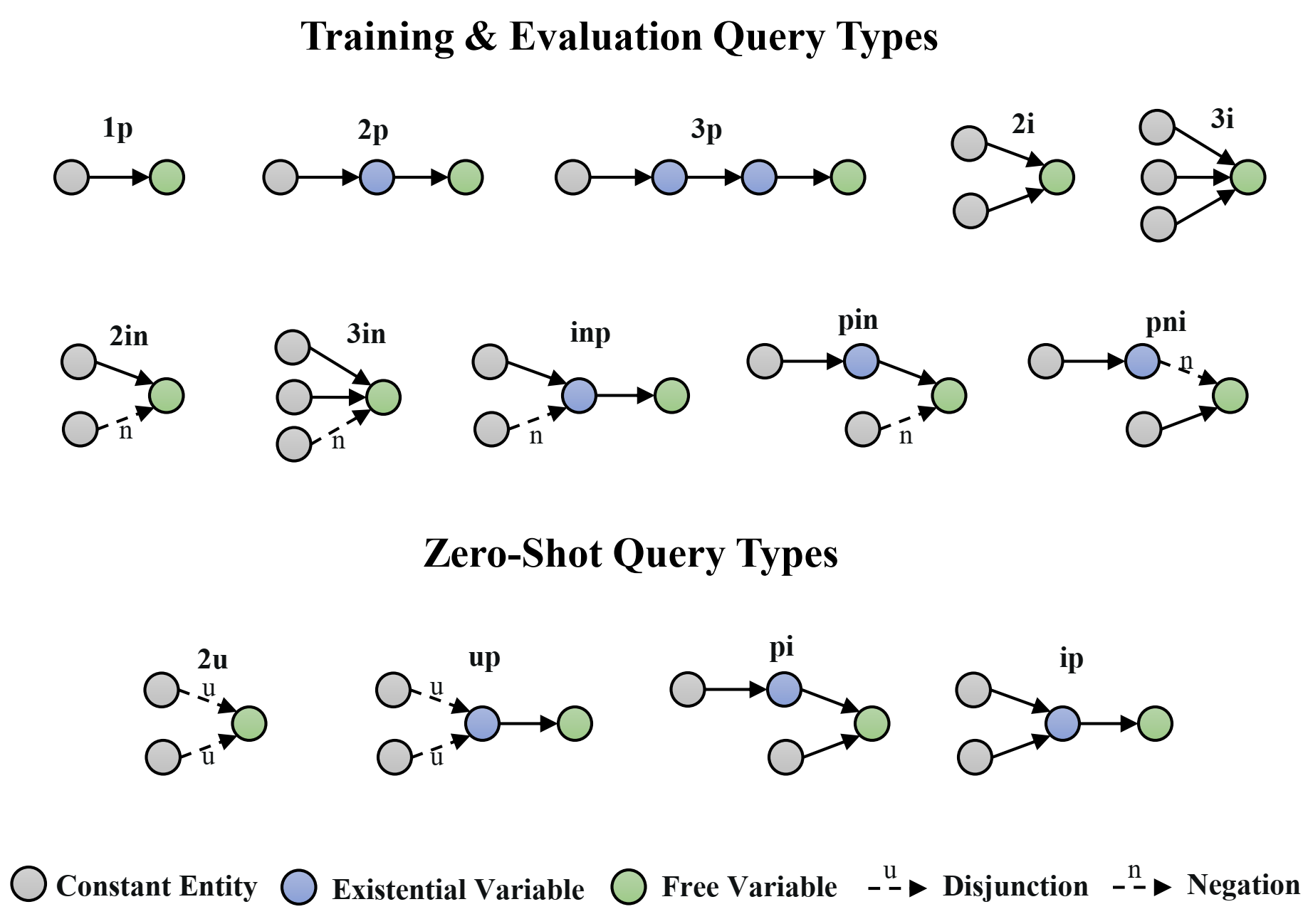}
    \caption{All the query types considered in our experiments, where $p$, $i$, $u$, and $n$ represent projection, intersection, union, and negation, respectively. The naming of each query type reflects how they were generated in the BetaE paper \cite{ren2020beta}.  }
    \label{figure3}
\end{figure}

\begin{table}[htbp]
  \centering
  \caption{Statistics of different query types used in the benchmark datasets. }
  \resizebox{0.9\linewidth}{!}{
    \begin{tabular}{ccccc}
    \toprule
    \textbf{Split} & \textbf{Query Types} & \textbf{FB15k} & \textbf{FB15k-237} & \textbf{NELL995} \\
    \midrule
    \multirow{2}[2]{*}{\textbf{Train}} & 1p, 2p, 3p, 2i, 3i & 273,710 & 149,689 & 107,982 \\
          & 2in, 3in, inp, pin, pni & 27,371 & 14,968 & 10,798 \\
    \midrule
    \multirow{2}[2]{*}{\textbf{Valid}} & 1p    & 59,078 & 20,094 & 16,910 \\
          & Others & 8,000 & 5,000 & 4,000 \\
    \midrule
    \multirow{2}[2]{*}{\textbf{Test}} & 1p    & 66,990 & 22,804 & 17,021 \\
          & Others & 8,000 & 5,000 & 4,000 \\
    \bottomrule
    \end{tabular}%
    }
  \label{stat}%
\end{table}%

\section{Experiments}

\subsection{Experimental Settings}

\subsubsection{Datasets and Queries}
\label{dataset}

We evaluate our method on three commonly-used standard knowledge graphs: FB15k \cite{bordes2013translating}, FB15k-237 \cite{toutanova2015observed}, and NELL995 \cite{xiong2017deeppath}. 
For a fair comparison with previous works, we use the datasets of complex queries proposed by BetaE \cite{ren2020beta}, 
which consist of five conjunctive query types ($1p/2p/3p/2i/3i$), five query types with atomic negations ($2in/3in/inp/pin/pni$), and four zero-shot query types ($pi/ip/2u/up$),
as illustrated in Figure \ref{figure3}. 
The datasets provided by BetaE introduce complex queries with hard answers that cannot be obtained by traversing the KG directly. 
That is, the evaluation focuses on discovering such hard answers to complex queries. 
We follow previous works \cite{ren2020beta, wang2023logical,zhang2021cone} and train our model with five conjunctive query types and five query types with atomic negations. 
When evaluating the model, we consider all query types including the four zero-shot query types. The statistics for each dataset are presented in Table \ref{stat}. 

\begin{table*}[htbp]
  \centering
  \caption{
  The MRR results of baselines and our model over three KGs. $\mathbf{Avg_{p}}$ represents the average score of EPFO queries and $\mathbf{Avg_{n}}$ denote the average score of queries with negation. The boldface indicates the best results, and the second-best ones are marked with \underline{underlines}. }
  \resizebox{0.88\linewidth}{!}{
    \begin{tabular}{cccccccccccccccccc}
    \toprule
    \textbf{Dataset} & \textbf{Model} & \textbf{1p} & \textbf{2p} & \textbf{3p} & \textbf{2i} & \textbf{3i} & \textbf{pi} & \textbf{ip} & \textbf{2u} & \textbf{up} & \textbf{2in} & \textbf{3in} & \textbf{inp} & \textbf{pin} & \textbf{pni} & \textbf{$\mathbf{Avg_{p}}$} & \textbf{$\mathbf{Avg_{n}}$} \\
    \midrule
    \multirow{10}[4]{*}{FB15k} & BetaE & 65.1  & 25.7  & 24.7  & 55.8  & 66.5  & 43.9  & 28.1  & 40.1  & 25.2  & 14.3  & 14.7  & 11.5  & 6.5   & 12.4  & 41.6  & 11.8 \\
          & ConE  & 73.3  & 33.8  & 29.2  & 64.4  & 73.7  & 50.9  & 35.7  & \underline{55.7}  & 31.4  & 17.9  & 18.7  & 12.5  & 9.8   & 15.1  & 49.8  & 14.8 \\
          & Q2P   & 82.6  & 30.8  & 25.5  & 65.1  & 74.7  & 49.5  & 34.9  & 32.1  & 26.2  & \underline{21.9}  & 20.8  & 12.5  & 8.9   & \underline{17.1}  & 46.8  & 16.4 \\
          & MLP   & 67.1  & 31.2  & 27.2  & 57.1  & 66.9  & 45.7  & 33.9  & 38.0    & 28.0    & 17.2  & 17.8  & 13.5  & 9.1   & 15.2  & 43.9  & 14.5 \\
          & GammaE & 76.5  & 36.9  & \underline{31.4}  & 65.4  & 75.1  & \underline{53.9}  & 39.7  & 53.5  & 30.9  & 20.1  & 20.5  & 13.5  & \underline{11.8}  & \underline{17.1}  & 51.3  & 16.6 \\
          & CylE  & 78.8  & 37.0    & 30.9  & 66.9  & 75.7  & 53.8  & 40.8  & \textbf{59.4}  & \underline{33.5}  & 15.7  & 16.3  & 13.7  & 7.8   & 13.9  & \underline{53.0}    & 13.5 \\
\cmidrule{2-18}          & \multicolumn{17}{l}{(Using pre-trained neural link predictor)} \\
          & CQD-CO & \textbf{89.4}  & 27.6  & 15.1  & 63.0    & 65.5  & 46.0    & 35.2  & 42.9  & 23.2  & 0.2   & 0.2   & 4.0     & 0.1   & \textbf{18.4}  & 45.3  & 4.6 \\
          & LMPNN & 85.0    & \underline{39.9}  & 28.6  & \underline{68.2}  & \underline{76.5}  & 46.7  & \underline{43.0}    & 36.7  & 31.4  & \textbf{29.1}  & \textbf{29.4}  & \underline{14.9}  & 10.2  & 16.4  & 50.6  & \textbf{20.0} \\
          & CLMPT & \underline{86.1}  & \textbf{43.3}  & \textbf{33.9}  & \textbf{69.1}  & \textbf{78.2}  & \textbf{55.1}  & \textbf{46.6}  & 46.1  & \textbf{37.3}  & 21.2  & \underline{22.9}  & \textbf{17.0}    & \textbf{13.0}    & 15.3  & \textbf{55.1} & \underline{17.9} \\
    \midrule
    \multirow{10}[4]{*}{FB15k-237} & BetaE & 39.0    & 10.9  & 10.0    & 28.8  & 42.5  & 22.4  & 12.6  & 12.4  & 9.7   & 5.1   & 7.9   & 7.4   & 3.6   & 3.4   & 20.9  & 5.4 \\
          & ConE  & 41.8  & 12.8  & \underline{11.0}    & 32.6  & 47.3  & 25.5  & 14.0    & \underline{14.5}  & 10.8  & 5.4   & 8.6   & 7.8   & 4.0     & 3.6   & 23.4  & 5.9 \\
          & Q2P   & 39.1  & 11.4  & 10.1  & 32.3  & 47.7  & 24.0    & 14.3  & 8.7   & 9.1   & 4.4   & 9.7   & 7.5   & 4.6   & 3.8   & 21.9  & 6.0 \\
          & MLP   & 42.7  & 12.4  & 10.6  & 31.7  & 43.9  & 24.2  & 14.9  & 13.7  & 9.7   & 6.6   & 10.7  & 8.1   & 4.7   & 4.4   & 22.6  & 6.9 \\
          & GammaE & 43.2  & 13.2  & \underline{11.0}    & 33.5  & 47.9  & \underline{27.2}  & 15.9  & 13.9  & 10.3  & 6.7   & 9.4   & \textbf{8.6}   & \underline{4.8}   & 4.4   & 24.0    & 6.8 \\
          & CylE  & 42.9  & \underline{13.3}  & \textbf{11.3}  & \underline{35.0}    & \underline{49.0}    & 27.0    & 15.7  & \textbf{15.3}  & \textbf{11.2}  & 4.9   & 8.3   & \underline{8.2}   & 3.7   & 3.4   & \underline{24.5} & 5.7 \\
\cmidrule{2-18}          & \multicolumn{17}{l}{(Using pre-trained neural link predictor)} \\
          & CQD-CO & \textbf{46.7}  & 10.3  & 6.5   & 23.1  & 29.8  & 22.1  & 16.3  & 14.2  & 8.9   & 0.2   & 0.2   & 2.1   & 0.1   & \textbf{6.1}   & 19.8  & 1.7 \\
          & LMPNN & \underline{45.9}  & 13.1  & 10.3  & 34.8  & 48.9  & 22.7  & \underline{17.6}  & 13.5  & 10.3  & \textbf{8.7}   & \underline{12.9}  & 7.7   & 4.6   & \underline{5.2}   & 24.1  & \underline{7.8} \\
          & CLMPT & 45.7  & \textbf{13.7}  & \textbf{11.3}  & \textbf{37.4}  & \textbf{52.0}    & \textbf{28.2}  & \textbf{19.0}    & 14.3  & \underline{11.1}  & \underline{7.7}   & \textbf{13.7}  & 8.0     & \textbf{5.0}     & 5.1   & \textbf{25.9} & \textbf{7.9} \\
    \midrule
    \multirow{10}[4]{*}{NELL995} & BetaE & 53.0    & 13.0    & 11.4  & 37.6  & 47.5  & 24.1  & 14.3  & 12.2  & 8.5   & 5.1   & 7.8   & 10.0    & 3.1   & 3.5   & 24.6  & 5.9 \\
          & ConE  & 53.1  & 16.1  & 13.9  & 40.0    & 50.8  & 26.3  & 17.5  & 15.3  & 11.3  & 5.7   & 8.1   & 10.8  & 3.5   & 3.9   & 27.2  & 6.4 \\
          & Q2P   & 56.5  & 15.2  & 12.5  & 35.8  & 48.7  & 22.6  & 16.1  & 11.1  & 10.4  & 5.1   & 7.4   & 10.2  & 3.3   & 3.4   & 25.5  & 6.0 \\
          & MLP   & 55.2  & 16.8  & 14.9  & 36.4  & 48.0    & 22.7  & 18.2  & 14.7  & 11.3  & 5.1   & 8.0     & 10.0    & 3.6   & 3.6   & 26.5  & 6.1 \\
          & GammaE & 55.1  & 17.3  & 14.2  & \textbf{41.9}  & 51.1  & 26.9  & 18.3  & 15.1  & 11.2  & 6.3   & \underline{8.7}   & 11.4  & \textbf{4.0}     & 4.5   & 27.9  & \underline{7.0} \\
          & CylE  & 56.5  & 17.5  & 15.6  & 41.4  & \underline{51.2}  & 27.2  & 19.6  & 15.7  & 12.3  & 5.6   & 7.5   & 11.2  & 3.4   & 3.7   & 28.5  & 6.3 \\
\cmidrule{2-18}          & \multicolumn{17}{l}{(Using pre-trained neural link predictor)} \\
          & CQD-CO & \textbf{60.8}  & \underline{18.3}  & 13.2  & 36.5  & 43.0    & 30.0    & 22.5  & \underline{17.6}  & 13.7  & 0.1   & 0.1   & 4.0     & 0.0     & \textbf{5.2}   & 28.4  & 1.9 \\
          & LMPNN & \underline{60.6}  & \textbf{22.1}  & \underline{17.5}  & 40.1  & 50.3  & \underline{28.4}  & \textbf{24.9}  & 17.2  & \underline{15.7}  & \textbf{8.5}   & \textbf{10.8}  & \textbf{12.2}  & \underline{3.9}   & \underline{4.8}   & \underline{30.7}  & \textbf{8.0} \\
          & CLMPT & 58.9  & \textbf{22.1}  & \textbf{18.4}  & \underline{41.8}  & \textbf{51.9}  & \textbf{28.8}  & \underline{24.4}  & \textbf{18.6}  & \textbf{16.2}  & \underline{6.6}   & 8.1   & \underline{11.8}  & 3.8   & 4.5   & \textbf{31.3} & \underline{7.0} \\
    \bottomrule
    \end{tabular}%
    }
  \label{main results}%
\end{table*}%

\subsubsection{Evaluation Protocol}
\label{evaluation protocol}

The evaluation scheme follows the previous works \cite{ren2020beta, wang2023logical}, which divides the answers to each complex query into easy and hard sets. 
For test and validation splits, we define hard answers as those that cannot be obtained by direct graph traversal on KG. 
In order to get these hard answers, the model is required to impute at least one missing edge, which means the model needs to complete non-trivial reasoning \cite{Ren*2020Query2box:,ren2020beta}. 
Specifically, for each hard answer of a query, we rank it against non-answer entities based on their cosine similarity with the free variable embedding of the query and calculate the Mean Reciprocal Rank (MRR).

\subsubsection{Baselines}
\label{baselines}

We consider the state-of-the-art neural complex query answering models for EFO-1 queries in recent years as our baselines to compare: BetaE \cite{ren2020beta}, ConE \cite{zhang2021cone}, Q2P \cite{bai2022query2particles}, MLP \cite{amayuelas2022neural}, GammaE \cite{yang2022gammae}, CylE \cite{nguyen2023cyle}, CQD-CO \cite{arakelyan2020complex}, and LMPNN \cite{wang2023logical}, 
where CQD-CO and LMPNN use pre-trained neural link predictor, while other models do not. 
In addition, some other neural CQA models cannot handle logical negation operators, such as BIQE \cite{kotnis2021answering}, PREM \cite{choudhary2021probabilistic}, kgTransformer \cite{liu2022mask}, etc., which use the EPFO queries generated by Q2B \cite{Ren*2020Query2box:} for training and evaluation. 
We compare these works on the Q2B datasets in Appendix \ref{app q2b}. 
We also compare our work with symbolic integration CQA models in Appendix \ref{app sym}.

\subsubsection{Model Details}
\label{Model Details}

Following LMPNN \cite{wang2023logical} and CQD-CO \cite{arakelyan2020complex}, we choose ComplEx-N3 \cite{lacroix2018canonical, trouillon2016complex} as the neural link predictor in our work and use the ComplEx-N3 checkpoints released by CQD-CO to conduct experiments. 
The rank of ComplEx-N3 is 1,000, and the epoch for the checkpoints is 100. 
For all datasets, we use a two-layer transformer encoder with eight self-attention heads, where the dimension of the hidden layer of the feed-forward network (FFN) in the transformer is 8,192. 
For the training objective, the negative sample size $K$ is 128, and $T$ is chosen as 0.05 for FB15k-237 and FB15k and 0.1 for NELL995. 
For more details about the implementation and experiments, please refer to Appendix \ref{more experimental details}.

\begin{table}[htbp]
  \centering
  \caption{The average MRR results of models with or without conditional logical message passing on answering EFO-1 queries. $\downarrow $Memory and $\downarrow $Time represent the GPU memory usage reduction and time savings with conditional logical message passing, respectively. The boldface indicates the best results. 
  }
  \resizebox{0.9\linewidth}{!}{
    \begin{tabular}{cccccc}
    \toprule
    \textbf{Dataset} & \textbf{Model} & \textbf{$\mathbf{Avg_{p}}$} & \textbf{$\mathbf{Avg_{n}}$} & \textbf{$\downarrow $Memory} & \textbf{$\downarrow $Time} \\
    \midrule
    \multirow{4}[4]{*}{FB15k-237} & LMPNN & 24.1  & \textbf{7.8}   & \multirow{2}[2]{*}{$\downarrow $1.2\%} & \multirow{2}[2]{*}{$\downarrow $5.0\%} \\
          & LMPNN w/ C & \textbf{24.3}  & \textbf{7.8}   &       &  \\
\cmidrule{2-6}          & CLMPT w/o C & 25.8  & 7.7   & \multirow{2}[2]{*}{$\downarrow $9.6\%} & \multirow{2}[2]{*}{$\downarrow $12.3\%} \\
          & CLMPT & \textbf{25.9}  & \textbf{7.9}   &       &  \\
    \midrule
    \multirow{4}[4]{*}{NELL995} & LMPNN & 30.7  & 8.0     & \multirow{2}[2]{*}{$\downarrow $0.6\%} & \multirow{2}[2]{*}{$\downarrow $8.3\%} \\
          & LMPNN w/ C & \textbf{31.0}    & \textbf{8.2}   &       &  \\
\cmidrule{2-6}          & CLMPT w/o C & \textbf{31.4}  & \textbf{7.0}     & \multirow{2}[2]{*}{$\downarrow $8.0\%} & \multirow{2}[2]{*}{$\downarrow $11.0\%} \\
          & CLMPT & 31.3  & \textbf{7.0}     &       &  \\
    \bottomrule
    \end{tabular}%
    }
  \label{CLMP}%
\end{table}%

\begin{table*}[htbp]
  \centering
  \caption{The MRR results of different hyperparameter settings and different pooling approaches. The best results are bold.}
  \resizebox{0.8\linewidth}{!}{
    \begin{tabular}{ccccccccccccccccc}
    \toprule
    \textbf{Model} & \textbf{1p} & \textbf{2p} & \textbf{3p} & \textbf{2i} & \textbf{3i} & \textbf{pi} & \textbf{ip} & \textbf{2u} & \textbf{up} & \textbf{2in} & \textbf{3in} & \textbf{inp} & \textbf{pin} & \textbf{pni} & \textbf{$\mathbf{Avg_{p}}$} & \textbf{$\mathbf{Avg_{n}}$} \\
    \midrule
    $TE$ layer = 1 & 45.8  & 13.2  & 10.9  & 37.3  & 51.9  & 27.7  & 18.2  & 14.6  & 10.9  & \textbf{7.7}   & 13.3  & 8.1   & 5.3   & 4.5   & 25.6  & 7.8 \\
    $TE$ layer = 3 & 45.8      & \textbf{13.9}      & \textbf{11.4}      & 37.3      & 51.8      & \textbf{29.0}      & \textbf{19.2}      & 14.4      & 11.2      & 7.6      & 13.1      & 8.1      & 5.0      & 4.9      & \textbf{26.0}      & 7.7 \\
    \midrule
    $d_{\mathrm {FFN}}$ = 4096 & 45.8  & 13.5  & 11.1  & 37.3  & 51.7  & 27.9  & 18.6  & 14.3  & 11.2  & \textbf{7.7}   & 13.2  & 8.1   & 5.1   & 4.9   & 25.7  & 7.8 \\
    \midrule
    self-attention head = 4 & 45.9  & 13.6  & 11.3  & 37.4  & \textbf{52.0}    & 28.1  & 18.7  & 14.6  & \textbf{11.3}  & 7.4   & 13.1  & 8.1   & 5.0     & 4.8   & 25.9  & 7.7 \\
    self-attention head = 16 & 45.0      & 12.8      & 10.3      & 36.4      & 50.4      & 26.8      & 18.0      & 14.5      & 10.6      & 7.5      & 12.9      & 7.7      & 5.2      & 4.3      & 25.0      & 7.5 \\
    \midrule
    $T$ = 0.01 & 45.8  & 12.9  & 10.3  & 35.1  & 48.9  & 26.4  & 16.8  & 14.1  & 10.3  & 7.1   & 10.8  & 7.4   & \textbf{5.6}   & 4.0     & 24.5  & 7.0 \\
    $T$ = 0.1 & 44.9  & 13.5  & 10.7  & 34.4  & 47.2  & 26.5  & 18.7  & \textbf{15.6}  & 11.1  & 7.6   & 10.2  & 8.1   & 4.2   & \textbf{5.8}   & 24.7  & 7.2 \\
    \midrule
    Max Pooling & \textbf{46.1}      & 13.8      & 11.1      & 36.8      & 50.5      & 26.0      & 17.5      & 14.4      & 11.1      & 5.1      & 9.5      & 7.3      & 3.7      & 3.4      & 25.2      & 5.8 \\
    Sum Pooling & 45.9  & 13.8  & 11.0    & \textbf{37.5}  & 51.9  & 28.5  & 18.7  & 14.3  & \textbf{11.3}  & 7.6   & 13.6  & \textbf{8.2}   & 5.0     & 4.9   & 25.9  & \textbf{7.9} \\
    \midrule
    Default Setting & 45.7  & 13.7  & 11.3  & 37.4  & \textbf{52.0}    & 28.2  & 19.0    & 14.3  & 11.1  & \textbf{7.7}   & \textbf{13.7}  & 8.0     & 5.0     & 5.1   & 25.9  & \textbf{7.9} \\
    \bottomrule
    \end{tabular}%
    }
  \label{hyperpara}%
\end{table*}%

\subsection{Main Results}

Table \ref{main results} shows the MRR results of CLMPT and neural CQA baselines about answering EFO-1 queries over three KGs. 
It is found that CLMPT reaches the best performance on average for EPFO queries across all three datasets. 
Compared with LMPNN, which is most relevant to our work, CLMPT achieves average performance improvements of 8.9\%, 7.5\%, and 2.0\% for EPFO queries on FB15k, FB15K-237, and NELL995, respectively. 
Since conditional logical message passing has little impact on the model performance (it will be later discussed in Section \ref{CLMP ana}), these performance improvements are mainly due to the transformer-based node embedding update scheme. 
This shows that this approach, which explicitly models dependencies between input elements by assigning adaptive weights to the messages and the corresponding node through the self-attention mechanism, is more suitable for the logical query graphs of EPFO queries than the GIN-like approach. 
For the negative queries, 
CLMPT achieves the best average performance on FB15k-237 and sub-optimal performance on the remaining two KGs. 
One potential explanation for these sub-optimal results is that the transformer architecture lacks inductive biases and relies more on the training data. 
As shown in the statistics in Table \ref{stat}, the number of negative queries in the training set is an order of magnitude less than the EPFO queries, 
which may lead to the tendency of the transformer encoder to learn the patterns relevant to answering EPFO queries, resulting in insufficient modeling of relevant patterns that answer negative queries.

\subsection{Ablation Study}

\subsubsection{Conditional Logical Meassage Passing}
\label{CLMP ana}

The logical message passing mechanism can be viewed as learning an embedding for the variable by utilizing information of constant entities. 
It makes this embedding similar to the embedding of the solution entity in the embedding space of the pre-trained neural link predictor. 
As in the previous analyses, we argue that for a constant node with the corresponding entity embedding, it is unnecessary to update its embedding in the forward passing of the model. 
To verify this, we conduct experiments on whether to pass messages to the constant node and update its embedding, and the results are shown in Table \ref{CLMP}, where "w/ C" indicates "with conditional logical message passing" and "w/o C" indicates "without conditional logical message passing". 
From the experimental results, the performance of LMPNN with conditional logical message passing on the two KGs is not reduced but improved. 
CLMPT achieves better results on FB15k-237 than the variant that does not use conditional logical message passing, only slightly underperforming the variant on the EPFO queries of NELL995. 
In addition, we evaluate how much unnecessary computational cost the conditional logical message passing mechanism can avoid on an RTX 3090 GPU. 
Specifically, we calculate the relative percentage of the reduction in GPU memory usage and the relative percentage of training time saved: 
\begin{equation}
\begin{split}
    \downarrow \mathrm {Memory} =(usage_{woC}-usage_{C})/usage_{woC}, \label{mem}
\end{split}
\end{equation}
\begin{equation}
\begin{split}
    \downarrow \mathrm {Time} =(t_{woC}-t_{C})/t_{woC}, \label{timesaveings}
\end{split}
\end{equation}
where $usage_{C}$ and $t_{C}$ respectively refer to the GPU memory usage and the required training time when the model uses conditional logical message passing, while $usage_{woC}$ and $t_{woC}$ correspond to the situation without conditional logical message passing. 
As shown in Table \ref{CLMP}, conditional logical message passing reduces computational costs without significantly negatively impacting model performance. 
In particular, 
the computational cost reduction is particularly significant for CLMPT, 
which uses the more complex transformer architecture. 
The above results confirm our view that the model does not need to consider the constant entity nodes in the query graph during the forward message passing.

\begin{table}[htbp]
  \centering
  \caption{The average MRR results for experiments on whether to freeze the Neural Link Predictor (NLP). The boldface indicates the best results. }
  \resizebox{0.8\linewidth}{!}{
    \begin{tabular}{cccccc}
    \toprule
    \multirow{2}[4]{*}{\textbf{Dataset}} & \multirow{2}[4]{*}{\textbf{Model}} & \multicolumn{2}{c}{\textbf{Frozen NLP}} & \multicolumn{2}{c}{\textbf{Trainable NLP}} \\
\cmidrule{3-6}          &       & \textbf{$\mathbf{Avg_{p}}$} & \textbf{$\mathbf{Avg_{n}}$} & \textbf{$\mathbf{Avg_{p}}$} & \textbf{$\mathbf{Avg_{n}}$} \\
    \midrule
    \multirow{2}[2]{*}{FB15k} & LMPNN & 50.6  & \textbf{20.0} & 51.5  & \textbf{18.9} \\
          & CLMPT & \textbf{54.2} & 17.3  & \textbf{55.1} & 17.9 \\
    \midrule
    \multirow{2}[2]{*}{FB15k-237} & LMPNN & 24.1  & \textbf{7.8} & 24.4  & 7.5 \\
          & CLMPT & \textbf{25.4} & 7.7   & \textbf{25.9} & \textbf{7.9} \\
    \midrule
    \multirow{2}[2]{*}{NELL995} & LMPNN & 30.7  & \textbf{8.0} & 30.1  & \textbf{7.9} \\
          & CLMPT & \textbf{32.0} & 6.8   & \textbf{31.3} & 7.0 \\
    \bottomrule
    \end{tabular}%
    }
  \label{freeze experiment}%
\end{table}%

\begin{table*}[htbp]
  \centering
  \caption{The MRR results of CLMPT-small using frozen neural link predictor and using trainable neural link predictor. $\mathbf{Param_{T}}$ and $\mathbf{Param_{F}}$ represent trainable and frozen parameters of the model, respectively. The F in parentheses indicates that the model uses a frozen neural link predictor, and T indicates that it uses a trainable one.}
  \resizebox{1\linewidth}{!}{
    \begin{tabular}{cccccccccccccccccccc}
    \toprule
    \textbf{Dataset} & \textbf{Model} & \textbf{1p} & \textbf{2p} & \textbf{3p} & \textbf{2i} & \textbf{3i} & \textbf{pi} & \textbf{ip} & \textbf{2u} & \textbf{up} & \textbf{2in} & \textbf{3in} & \textbf{inp} & \textbf{pin} & \textbf{pni} & \textbf{$\mathbf{Avg_{p}}$} & \textbf{$\mathbf{Avg_{n}}$} & \textbf{$\mathbf{Param_{T}}$} & \textbf{$\mathbf{Param_{F}}$} \\
    \midrule
    \multirow{3}[2]{*}{FB15k-237} & LMPNN & 45.9  & 13.1  & 10.3  & 34.8  & 48.9  & 22.7  & 17.6  & 13.5  & 10.3  & 8.7   & 12.9  & 7.7   & 4.6   & 5.2   & 24.1  & \textbf{7.8}   & 16M   & 30M \\
          & CLMPT-small(F) & 46.1  & 12.6  & 10.0    & 36.1  & 50.4  & 26.7  & 17.5  & 14.9  & 10.3  & 7.7   & 11.6  & 7.5   & 4.4   & 5.0     & 25.0    & 7.3   & 32M   & 30M \\
          & CLMPT-small(T) & 45.7  & 13.2  & 11.0    & 37.2  & 51.6  & 27.4  & 18.2  & 14.3  & 11.1  & 7.4   & 13.1  & 8.1   & 5.1   & 4.7   & \textbf{25.5}  & 7.7   & 62M   & 0M \\
    \midrule
    \multirow{3}[2]{*}{NELL995} & LMPNN & 60.6  & 22.1  & 17.5  & 40.1  & 50.3  & 28.4  & 24.9  & 17.2  & 15.7  & 8.5   & 10.8  & 12.2  & 3.9   & 4.8   & 30.7  & \textbf{8.0}     & 33M   & 128M \\
          & CLMPT-small(F) & 60.6  & 21.7  & 17.7  & 42.2  & 51.7  & 30.7  & 24.5  & 19.4  & 15.6  & 6.4   & 7.9   & 11.0    & 3.7   & 4.5   & \textbf{31.6}  & 6.7   & 32M   & 128M \\
          & CLMPT-small(T) & 59.1  & 21.3  & 17.5  & 41.8  & 51.8  & 29.0    & 23.5  & 19.0    & 15.6  & 6.4   & 7.9   & 11.2  & 3.8   & 4.5   & 31.0    & 6.8   & 160M  & 0M \\
    \bottomrule
    \end{tabular}%
    }
  \label{parameter experiment}%
\end{table*}%

\subsubsection{Hyperparamaters and Pooling Approaches}

We evaluate the performance of CLMPT with different hyperparameter settings and pooling approaches on FB15k-237. 
The default setting we adopted is described in Section \ref{Model Details}. 
As is shown in Table \ref{hyperpara}, regardless of reducing the number of transformer encoder layers or halving the dimensions of the hidden layer of the FFN, the model still achieves the best performance than the baselines on FB15k-237, which reflects the effectiveness of our proposed method. 
When the number of transformer encoder layers increases, 
the performance changes little, meaning the two-layer transformer encoder is sufficient. 
In addition, we can find that $T$ in NCE loss is very important to the performance. 
When $T$ is 0.01 or 0.1, the performance of the model is still competitive on EPFO queries, but there is a large gap with the average performance under the default settings. 
For the pooling approaches, we evaluate max pooling and sum pooling. 
The experimental results show max pooling does not perform as well as sum pooling and mean pooling.

\subsubsection{Frozen or Trainable Neural Link Predictors}

For the pre-trained neural link predictor, it can be frozen or trainable. 
As mentioned above (see Section \ref{4.4}), in our work, 
we optimize all parameters in the pre-trained neural link predictor by default. 
It is worth noting that the pre-trained neural link predictor is frozen in LMPNN. 
In order to explore the impact of freezing the pre-trained neural link predictor on model performance, 
we conduct experiments on three KGs, and the results are shown in Table \ref{freeze experiment}.

For LMPNN, using frozen neural link predictors performs better than using trainable ones, except for EPFO queries on FB15k and FB15k-237. 
Therefore, in the main results, we only compare the MRR results of LMPNN using frozen neural link predictors, i.e. the results reported in the original paper \cite{wang2023logical}. 
For CLMPT, with the exception of EPFO queries on NELL995, CLMPT using the trainable neural link predictor performs relatively better. 
Thus, we continue to optimize the pre-trained neural link predictor on complex queries by default. 
For these results of CLMPT, 
according to the progress of pre-training and fine-tuning paradigms \cite{devlin2018bert, liu2019roberta}, 
one possible reason is that the neural link predictor is pre-trained on one-hop queries, and tuning its parameters on complex queries can make the embeddings of entities and relations more suitable for the patterns and characteristics of CQA. 
The better performance of LMPNN using trainable neural link predictors on EPFO queries of FB15k and FB15K-237 can also support our argument to some extent because the number of EPFO queries in the datasets is much more than negative ones; in this case, the model is more inclined to learn patterns that answer EPFO queries.

Although CLMPT, which uses the trainable neural link predictor, has more training parameters, it still significantly outperforms LMPNN on EPFO queries across all datasets when the neural link predictor is frozen. This reflects the effectiveness of CLMPT.

\subsection{Analyses on Model Parameters}
\label{app param}

Since CLMPT introduces the transformer, 
it has more parameters than LMPNN. 
To further evaluate the effectiveness of CLMPT, 
we compare LMPNN using a smaller model whose parameters are close to those of LMPNN. 
Specifically, we set $TE\ \mathrm {layer}  = 1$ and $d_{\mathrm{FFN}}=4096$, and we represent such a smaller model as CLMPT-small. 
Since the parameters of the neural link predictor in LMPNN are frozen, 
we consider using both frozen and trainable neural link predictors for fair comparison. 
We represent the models in these two cases as CLMPT-small(F) and CLMPT-small(T), respectively. 
The experimental results are shown in Table \ref{parameter experiment}. 
When their model parameters are close, 
CLMPT still achieves the best average performance on EPFO queries. 
In the case of freezing the neural link predictor, 
CLMPT-small(F) has fewer trainable parameters than LMPNN on NELL995, 
but it still significantly outperforms LMPNN on average in anwsering EPFO queries. 
These experimental results reflect the effectiveness of CLMPT.

\section{Conclusion}

In this paper, we propose CLMPT, a special message passing neural network, to answer complex queries over KGs. 
Based on one-hop inference by pre-trained neural link predictor on atomic formulas, CLMPT performs logical message passing conditionally on the node type. In the ablation study, we verify that this conditional message passing mechanism can effectively reduce the computational costs and even improve the performance of the model to some extent. 
Furthermore,  CLMPT uses the transformer to aggregate messages and update the corresponding node embedding. 
Through the self-attention mechanism, CLMPT can explicitly model logical dependencies between messages received by a node and between messages and that node by assigning adaptive weights to the messages and the node. 
The experimental results show that CLMPT achieves a strong performance through this transformer-based node embedding update scheme. 
Future work may integrate symbolic information into CLMPT to improve the performance in answering negative queries. 
For limitations, please refer to Appendix \ref{limitations}.

\begin{acks}
We thank the anonymous reviewers for their helpful feedbacks. 
The work described in this paper was partially funded by the National Natural Science Foundation of China (Grant Nos. 62272173, 62273109), the Natural Science Foundation of Guangdong Province (Grant Nos. 2024A1515010089, 2022A1515010179), the Science and Technology Planning Project of Guangdong Province (Grant No. 2023A0505050106), and the National Key R\&D Program of China (Grant No. 2023YFA1011601).
\end{acks}


\bibliography{references}
\balance
\bibliographystyle{ACM-Reference-Format}

\appendix

\section{More Experimental Details}
\label{more experimental details}

Our code is implemented using PyTorch. 
We use NVIDIA Tesla A10 GPU (24GB) and NVIDIA GeForce RTX 3090 GPU (24GB) to conduct all of our experiments. 
We use the AdamW \cite{loshchilov2018decoupled} optimizer, whose weight decay is 1e-4, to tune the parameters. 
The learning rate is 5e-5 for NELL995 and FB15k-237 and 1e-4 for FB15k. The batch size is 512 for NELL995 and FB15k-237 and 1,024 for FB15k. The training epoch for all datasets is 100. 
Following CQD-CO \cite{arakelyan2020complex} and LMPNN \cite{wang2023logical}, 
we choose ComplEx-N3 \cite{lacroix2018canonical, trouillon2016complex} as the neural link predictor for fair comparison. 
In this case, we consider the logical message encoding function $\rho$ corresponding to Equations \ref{rho11}--\ref{rho41}. 
For these equations, $\lambda$ is a hyperparameter that needs to be determined. 
In LMPNN application, for simplicity, LMPNN just lets $3\lambda \left \| \cdot \right \| = 1$ and then all denominators in these closed-form expressions are 1, namely: 
\begin{equation}
\begin{split}
    \rho (t,r,t\to h,0) := \overline{r}\otimes t,  \label{app rho1}
\end{split}
\end{equation} 
\begin{equation}
\begin{split}
    \rho (h,r,h\to t,0) := r \otimes h,  \label{app rho2}
\end{split}
\end{equation}
\begin{equation}
\begin{split}
    \rho (t,r,t\to h,1) := -\overline{r}\otimes t,  \label{app rho3}
\end{split}
\end{equation}
\begin{equation}
\begin{split}
    \rho (h,r,h\to t,1) := -r \otimes h.  \label{app rho4}
\end{split}
\end{equation}
To make a fair comparison with LMPNN, in CLMPT application, we also use the logical message encoding function $\rho$ defined in Equations \ref{app rho1}--\ref{app rho4} to compute messages passed to the variable nodes. 
For the experiments (see Section \ref{CLMP ana}) to evaluate how much unnecessary computational costs can be avoided by conditional logical message passing mechanism, we repeat the experiments three times on an RTX 3090 GPU and then average the calculated $\downarrow \mathrm {Memory}$ (see Equation \ref{mem}) and $\downarrow \mathrm {Time}$ (see Equation \ref{timesaveings}) metrics. 

\begin{table}[htbp]
  \centering
  \caption{The Hits@3 results of other neural CQA baselines on answering EPFO queries generated by Q2B \cite{Ren*2020Query2box:}. Avg represents the average results of all query types. The boldface indicates the best results. }
  \resizebox{1\linewidth}{!}{
    \begin{tabular}{cccccccccccc}
    \toprule
    \textbf{Dataset} & \textbf{Model} & \textbf{1p} & \textbf{2p} & \textbf{3p} & \textbf{2i} & \textbf{3i} & \textbf{pi} & \textbf{ip} & \textbf{2u} & \textbf{up} & \textbf{Avg} \\
    \midrule
    \multirow{7}[2]{*}{FB237-Q2B} & GQE   & 40.5  & 21.3  & 15.3  & 29.8  & 41.1  & 18.2  & 8.5   & 16.7  & 16.0    & 23.0 \\
          & Q2B   & 46.7  & 24.0    & 18.6  & 32.4  & 45.3  & 20.5  & 10.8  & 23.9  & 19.3  & 26.8 \\
          & BIQE  & 47.2  & 29.3  & 24.5  & 34.5  & 47.3  & 21.1  & 9.6   & 7.5   & 14.5  & 26.2 \\
          & PERM  & \textbf{52.0} & 28.6  & 21.6  & 36.1  & 49.0    & 21.2  & 12.8  & \textbf{30.5}  & 23.9  & 30.6 \\
          & kgTransformer & 45.9  & 31.2  & \textbf{27.6} & \textbf{39.8} & \textbf{52.8} & \textbf{28.6} & \textbf{18.9} & 26.3  & 21.4  & 32.5 \\
          & SILR  & 47.1  & 30.2  & 24.9  & 35.8  & 48.4  & 22.2  & 11.3  & 28.3  & 18.1  & 29.6 \\
          & CLMPT & 50.5  & \textbf{31.3} & 26.7  & 39.0    & 52.1  & 27.7  & 16.3  & 28.6 & \textbf{24.1} & \textbf{32.9} \\
    \midrule
    \multirow{7}[2]{*}{NELL-Q2B} & GQE   & 41.7  & 23.1  & 20.3  & 31.8  & 45.4  & 18.8  & 8.1   & 20.0    & 13.9  & 24.8 \\
          & Q2B   & 55.5  & 26.6  & 23.3  & 34.3  & 48.0    & 21.2  & 13.2  & 36.9  & 16.3  & 30.6 \\
          & BIQE  & 63.2  & 31.0    & 33.2  & 37.0    & 52.5  & 16.4  & 9.1   & 17.3  & 18.4  & 30.9 \\
          & PERM  & 58.1  & 28.6  & 24.3  & 35.2  & 50.8  & 19.5  & 14.3  & 46.0    & 20.0    & 32.8 \\
          & kgTransformer & 62.5  & \textbf{40.1} & \textbf{36.7} & 40.5  & \textbf{54.6} & \textbf{30.6} & 20.3  & 46.9  & 27.0    & 39.9 \\
          & SILR  & 64.1  & 32.9  & 33.7  & 37.6  & 53.2  & 17.7  & 5.5   & 45.0    & 18.0    & 34.2 \\
          & CLMPT & \textbf{64.3} & 39.9  & 36.5  & \textbf{42.1} & 54.1  & 28.1  & \textbf{22.1} & \textbf{48.9} & \textbf{33.4} & \textbf{41.1} \\
    \bottomrule
    \end{tabular}%
    }
  \label{hits3}%
\end{table}%

\begin{table}[htbp]
  \centering
  \caption{The MRR results of symbolic integration models and CLMPT. The boldface indicates the best results. }
  \resizebox{1\linewidth}{!}{
    \begin{tabular}{cccccccccccccccccc}
    \toprule
    \textbf{Dataset} & \textbf{Model} & \textbf{1p} & \textbf{2p} & \textbf{3p} & \textbf{2i} & \textbf{3i} & \textbf{pi} & \textbf{ip} & \textbf{2u} & \textbf{up} & \textbf{2in} & \textbf{3in} & \textbf{inp} & \textbf{pin} & \textbf{pni} & \textbf{$\mathbf{Avg_{p}}$} & \textbf{$\mathbf{Avg_{n}}$} \\
    \midrule
    \multirow{4}[4]{*}{FB15k-237} & GNN-QE & 42.8  & 14.7  & 11.8  & 38.3  & 54.1  & 31.1  & 18.9  & 16.2  & 13.4  & 10.0    & 16.8  & 9.3   & 7.2   & 7.8   & \textbf{26.8}  & \textbf{10.2} \\
          & ENeSy & 44.7  & 11.7  & 8.6   & 34.8  & 50.4  & 27.6  & 19.7  & 14.2  & 8.4   & 10.1  & 10.4  & 7.6   & 6.1   & 8.1   & 24.5  & 8.5 \\
\cmidrule{2-18}          & CLMPT-small(F) & 46.1  & 12.6  & 10.0    & 36.1  & 50.4  & 26.7  & 17.5  & 14.9  & 10.3  & 7.7   & 11.6  & 7.5   & 4.4   & 5.0     & 25.0    & 7.3 \\
          & CLMPT & 45.7  & 13.7  & 11.3  & 37.4  & 52.0    & 28.2  & 19.0    & 14.3  & 11.1  & 7.7   & 13.7  & 8.0     & 5.0     & 5.1   & 25.9  & 7.9 \\
    \midrule
    \multirow{4}[4]{*}{NELL995} & GNN-QE & 53.3  & 18.9  & 14.9  & 42.4  & 52.5  & 30.8  & 18.9  & 15.9  & 12.6  & 9.9   & 14.6  & 11.4  & 6.3   & 6.3   & 28.9  & 9.7 \\
          & ENeSy & 59.0    & 18.0    & 14.0    & 39.6  & 49.8  & 29.8  & 24.8  & 16.4  & 13.1  & 11.3  & 8.5   & 11.6  & 8.6   & 8.8   & 29.4  & \textbf{9.8} \\
\cmidrule{2-18}          & CLMPT-small(F) & 60.6  & 21.7  & 17.7  & 42.2  & 51.7  & 30.7  & 24.5  & 19.4  & 15.6  & 6.4   & 7.9   & 11.0    & 3.7   & 4.5   & \textbf{31.6}  & 6.7 \\
          & CLMPT & 58.9  & 22.1  & 18.4  & 41.8  & 51.9  & 28.8  & 24.4  & 18.6  & 16.2  & 6.6   & 8.1   & 11.8  & 3.8   & 4.5   & 31.3  & 7.0 \\
    \bottomrule
    \end{tabular}%
    }
  \label{sym experiment}%
\end{table}%

\section{Comparison with More Neural CQA Models on Q2B Datasets}
\label{app q2b}

To further evaluate the performance of CLMPT, 
we also consider comparing neural CQA models that cannot handle logical negation and are trained and evaluated on Q2B datasets \cite{Ren*2020Query2box:}, including GQE \cite{hamilton2018embedding}, Q2B \cite{Ren*2020Query2box:}, 
BIQE \cite{kotnis2021answering}, PERM \cite{choudhary2021probabilistic}, kgTransformer \cite{liu2022mask}, and SILR \cite{wang2023query}. 
Among these models, BIQE, kgTransformer, and SILR all use the transformer \cite{vaswani2017attention} to encode complex queries directly. 
BIQE and SILR serialize the query graph and treat the CQA task as the sequence learning task to solve. 
kgTransformer uses the pre-training and fine-tuning strategies based on Heterogeneous Graph Transformer (HGT) \cite{hu2020heterogeneous} to answer complex queries. 
In essence, kgTransformer is also a transformer-based graph neural network. 
We use the results reported in these papers \cite{liu2022mask, wang2023query} for comparison. 
Since they report the Hits@3 results, for a fair comparison, we also use Hits@3 as the evaluation metric, which calculates the proportion of correct answer entities ranked among the top 3. 

As shown in the results in Table \ref{hits3}, 
CLMPT reaches the best performance on average across all datasets. 
For those models that use the transformer to encode the entire query graph directly, namely BIQE, SILR, and kgTransformer, we speculate that their insufficient graph inductive biases may negatively affect their performance. 
To encode the entire query graph at once using the transformer, BIQE designs a special positional encoding scheme, and SILR designs a particular graph serialization method and structural prompt. 
These designs can be viewed as introducing specific graph inductive biases into the transformer to encode the query graph. 
For kgTransformer, it relies primarily on the graph inductive biases of the HGT architecture. 
According to the progress of the graph transformers \cite{DBLP:conf/icml/Ma0LRDCTL23, ying2021transformers}, graph inductive biases are critical to using the transformer to encode graph data. 
In particular, structural encoding is important for the graph transformers without message passing. 
Therefore, we speculate that the graph inductive biases introduced by these CQA models may not be sufficient to encode the query graph that defines complex logical dependencies. 
On the other hand, it is not clear how to introduce appropriate graph inductive biases for query graphs that define first-order logical dependencies. 
By contrast, CLMPT, based on logical message passing, uses the transformer to aggregate messages and update node embeddings rather than to encode the entire query graph. 
This difference makes CLMPT less dependent on graph inductive biases than the transformer-based CQA models mentioned above. 
However, it also suggests that appropriate graph inductive biases may enhance the performance of CLMPT. 
This extension can be explored in the future.

\section{Comparison with Symbolic Integration Models}
\label{app sym}

We discuss the differences between symbolic integration CQA models and neural CQA models in Section \ref{relate work sym}. 
The model proposed in this paper is a neural CQA model, 
so our comparisons and related discussions are mainly carried out within the scope of neural CQA models. 
Since symbolic information can enhance the performance of neural CQA models \cite{wang2023logical}, 
in this section, 
we consider two representative symbolic integration models, GNN-QE \cite{zhu2022neural} and ENeSy \cite{xu2022neural}, for comparison with CLMPT to show the potential. 
We list the MRR results reported in their original papers. 
The results are shown in Table \ref{sym experiment}. 
It is found that CLMPT is also competitive even with the symbolic integration models on answering EPFO queries. 
Specifically, 
CLMPT achieves a better average performance than these symbolic integration models on EPFO queries for NELL995. 
For FB15k-237, CLMPT still has a gap in performance with GNN-QE, but it can outperform ENeSy on EPFO queries. 
For the negative queries, 
there are still gaps between the neural CQA models and these symbolic integration models because the fuzzy sets equipped with symbolic information used in the reasoning process of these models can effectively deal with logical negation through probabilistic values. 
These results suggest that neural models can be potentially improved with symbolic
integration. 
However, it is worth noting that introducing additional symbolic information requires larger computational costs.

According to \cite{ren2023neural, wang2023logical}, 
for larger knowledge graphs, neural CQA models and symbolic integration models have different scalabilities. 
As we discussed in Section \ref{relate work sym},
symbolic integration models require more computing resources and are more likely to suffer from scalability issues.
For example, GNN-QE employs NBFNet \cite{zhu2021neural} to compute message passing on the whole KG, resulting in complexity that is linear to $(|\mathcal{E}|+|\mathcal{V}|)d$, where $|\mathcal{E}|$ is the number of edges in KG, $|\mathcal{V}|$ is the number of nodes in KG, and $d$ is the embedding dimension. 
For CLMPT, which only performs message passing on the query graph, 
the complexity is just $O(d)$. 
Besides, GNN-QE estimates the probability of whether each entity is the answer at each intermediate step, making the size of its fuzzy sets scale linearly with $|\mathcal{V}|$. 
As a result, GNN-QE requires much more computational costs, requiring 128GB GPU memory to run a batch size of 32. 
For CLMPT-small(F), which has only 32M trainable parameters, 
it only requires less than 12GB GPU memory to run a batch size of 1024. 
Even in this case, CLMPT-small(F) can still achieve performance competitive with GNN-QE on NELL995. 
We suspect that integrating symbolic information into CLMPT can improve the performance, especially on negative queries. 
However, exploring how to integrate symbolic information into CLMPT is beyond the scope of this paper. 
This extension is left for future work.

\section{Limitations}
\label{limitations}

Due to the use of the transformer architecture in the node embedding update scheme in CLMPT, more computational costs are required compared to some previous works based on multi-layer perceptron. 
In addition, CLMPT only achieves sub-optimal performance on negative queries. 
We suspect that integrating symbolic information into CLMPT may improve the performance on negative queries.

\end{sloppypar}
\end{document}